\documentclass[11pt,a4paper]{article}
\usepackage{authblk}
\usepackage[hyperref]{emnlp2020}
\usepackage{times}
\usepackage{latexsym}
\usepackage{csquotes}
\usepackage[fleqn]{amsmath}
\usepackage[11pt]{moresize}
\usepackage{anyfontsize}
\usepackage{setspace}
\usepackage{multirow}
\usepackage{subfig}
\usepackage{adjustbox}
\usepackage{amssymb}
\newcommand{\cmark}{\ding{51}}%
\newcommand{\xmark}{\ding{55}}%

\usepackage{xcolor}
\usepackage{mdframed}
\usepackage{makecell}
\usepackage{graphicx}
\usepackage{float}
\usepackage{tabu}
\usepackage{pifont}
\usepackage{latexsym}

\usepackage{amsmath}
\DeclareMathOperator*{\argmax}{arg\,max}

\usepackage{microtype}

\aclfinalcopy 
\def\aclpaperid{525} 

\title{ZEST: Zero-shot Learning from Text Descriptions\\
using Textual Similarity and Visual Summarization}
\author[1]{\textbf{Tzuf Paz-Argaman}}
\author[2]{\textbf{Yuval Atzmon}}
\author[1,2]{\textbf{Gal Chechik}}
\author[1,3]{\textbf{Reut Tsarfaty}}
\affil[1]{Bar-Ilan University}
\affil[2]{NVIDIA Research}
\affil[3]{Allen Institute for Artificial Intelligence}
\affil[ ]{\tt \{tzufar,reut.tsarfaty\}@gmail.com}

\date{}

\begin{document}
\maketitle

\begin{abstract}

We study the problem of recognizing visual entities from the textual descriptions of their classes. Specifically, given birds' images with free-text descriptions of their species, we learn to classify images of previously-unseen species based on specie descriptions. This setup has been studied in the vision community under the name {\em zero-shot learning from text}, focusing on learning to transfer knowledge about visual aspects of birds from seen classes to previously-unseen ones. Here, we suggest focusing on the textual description and distilling from the description the most relevant information to effectively match visual features to the parts of the text that discuss them. 
Specifically, (1) we propose to leverage the {\em similarity} between species, reflected in the similarity between text descriptions of the species. (2) we derive {\em visual summaries} of the texts, i.e., 
extractive summaries that focus on the {\em visual} features that tend to be reflected in images.
We propose a simple attention-based model augmented with the similarity and visual summaries components. Our empirical results consistently and significantly outperform the state-of-the-art on the largest benchmarks for {\em text-based zero-shot learning}, illustrating the critical importance of texts for zero-shot image-recognition.

\end{abstract}

\section{Introduction}

In computer vision, {\em zero shot-learning} (ZSL) for image classification is the problem of classifying images given auxiliary information. An image classification model is trained to classify images from a pre-defined set of classes. At test time, images from new classes are given, and the task is to transfer knowledge learned from seen classes during training to unseen test classes.

\begin{figure}[H]
\centering
\scalebox{0.48}{
\includegraphics[width=\textwidth]{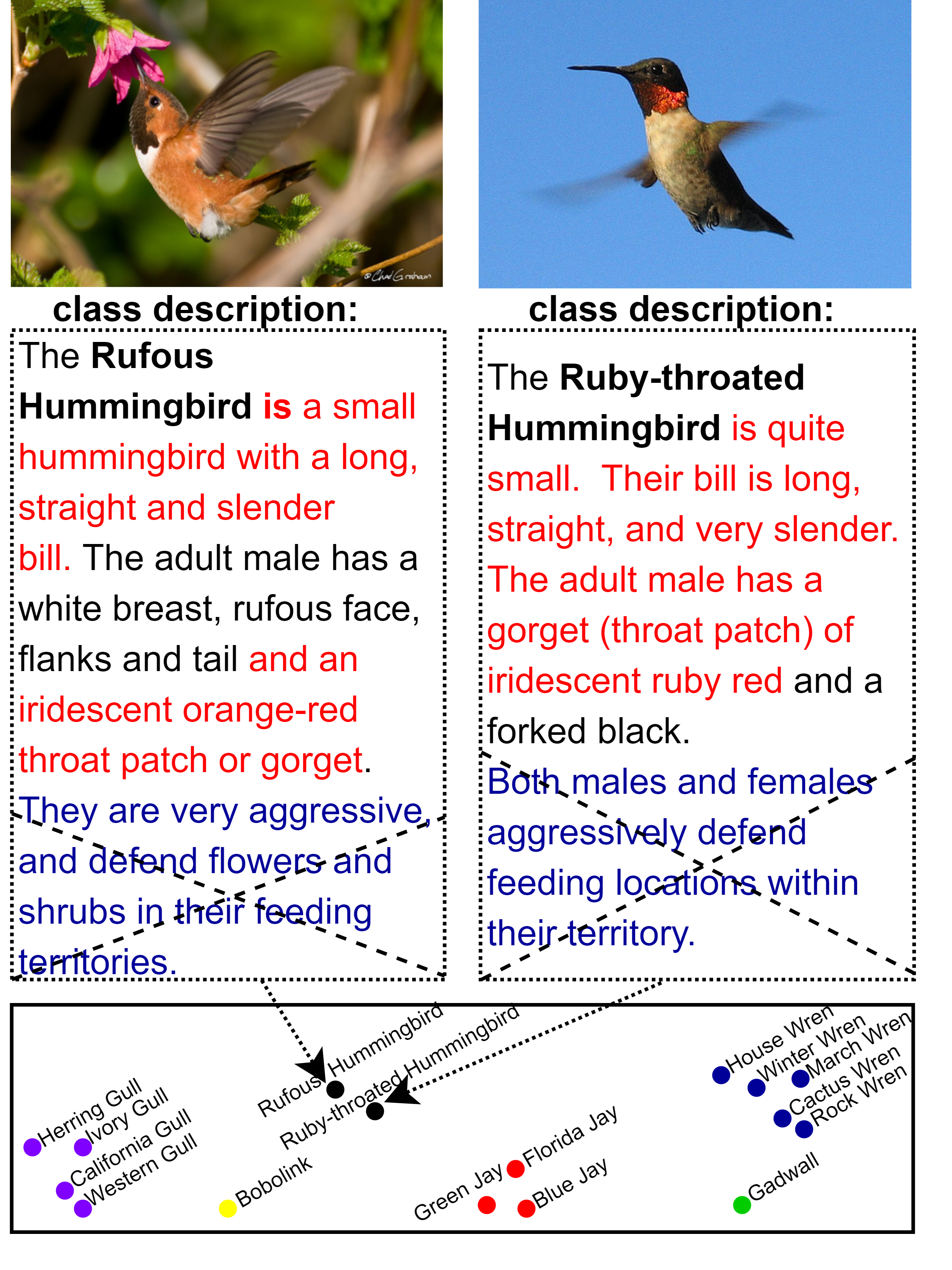}}
 \caption{An illustration of textual similarity and visually relevant descriptions in Wikipedia articles: (1) we aim to leverage the similarity within  texts (red) via document clustering (bottom box);  (2) we aim to extract similar (red) and dissimilar (black) visual descriptions, and remove non-visually relevant (blue) one. 
 } 
\label{fig:bird_example}%
\end{figure}

\vspace{-0.2in}

A common setup for ZSL assumes that the auxiliary information is a set of semantically meaningful properties (called attributes) describing the class (e.g., black-beak, long-tail) \citep{wah2011caltech,farhadi2009describing}. A different ZSL setup uses image captions as auxiliary information \citep{reed2016learning,Felix_2018_ECCV}. Typically, this auxiliary information is manually collected by human raters for each image (test and train alike) and averaged across images.
A more realistic approach 
relies on available online text descriptions of classes (e.g., Wikipedia) \cite{elhoseiny2017link}. It avoids expensive annotation and exposure to test images.

In this work, we classify bird species according to Wikipedia descriptions.
This task raises many challenges: 
(1) Differences between the birds are very small, which makes it a fine-grained classification task; (2) This is an expert task, 
and the text contains terminology that is unlikely to be familiar to a layman; and, on top of that
(3) The text descriptions of the classes are long, containing few visually relevant sentences.

As opposed to previous work on text-based ZSL employing textual descriptions \citep{zhu2018generative,elhoseiny2017link} that focused on the {\em visual} modality, here we focus on the {\em text} modality, and address a key question in ZSL: \textit {How can we identify text components that are visual in nature?}

To get an intuition about the task setup and our proposed solution, consider the following situation.
Imagine you have never seen a zebra but have seen a horse. What if you were given a text describing a zebra: \enquote{Zebras have hooves, mane, tail, pointed ears, and white and black stripes}. This description would probably be very close to a description of a horse having \enquote{hooves, mane, tail, pointed ears} and you would probably be looking for an image that reminds you of a horse but has \enquote{white and black stripes}. So, even without ever seeing a zebra, using text-descriptions of the zebra and knowledge already acquired about horses, one can correctly classify unknown classes like a zebra.

Our proposed solution has two-phases. First,
based on the intuition that similar objects (or images thereof) tend to have similar texts, we encode a similarity feature that enhances text descriptions' separability. 
In addition, we leverage the intuition that the differences between text descriptions of species would be their most salient visual features, and extract visually relevant descriptions from the text.

Our experiments empirically demonstrate both the {\em efficacy} and {\em generalization} capacity of our proposed solution. 
On two large ZSL datasets, in both the {\em easy} and {\em hard} scenarios, the similarity method obtains a ratio improvement of up to 18.3\%. With the addition of extracting visually relevant descriptions, we obtain a ratio improvement of up to 48.16\% over the state-of-the-art.
We further show that our visual-summarization method generalizes from the CUB dataset \citep{wah2011caltech} to the NAB dataset \citep{van2015building}, and we demonstrate its contribution to additional models by a ratio improvement of up to 59.62\%.

The contributions of this paper are threefold.
First, to the best of our knowledge, we are the first to showcase the critical importance of the text representation in zero-shot image-recognition scenarios, and we present two concrete text-based processing methods that vastly improve the results.
Second, we demonstrate the efficacy and generalizability of our proposed methods by applying them to both the {\em zero-shot} and {\em generalized zero-shot} tasks, outperforming all previously reported results on the CUB and NAB Benchmarks.  
Finally, we show that visual aspects learned from one dataset can be transferred effectively to another dataset without the need to obtain dataset-specific captions. 
The efficacy of our proposed solution on these benchmarks illustrates that purposefully exposing the {\em visual features in texts} is indispensable for tasks that learn to align the vision-and-language modalities.

\section{Background and Related Work}

Zero-shot learning (ZSL) aims at overcoming the need to label massive datasets for new categories, by learning the connections between images and prior auxiliary knowledge about their classes. At test-time, this auxiliary information compensates for the lack of previously-attained visual information about the new categories.
\par
Text-based ZSL is a specific multimodal instantiation of this learning task that uses natural language descriptions as the auxiliary information. Models for text-based ZSL are typically composed of three parts: (1) the text representation; (2) the image representation; (3) a compatibility function between the two.  While most previous work focused mainly on the latter two components, here we focus on the text. \par

Most ZSL studies for object recognition are aimed at processing the image modality. For example, \citet{xu2018attngan,lei2015predicting,qiao2016less,akata2016multi} rely on visual features extracted using 
Convolutional Neural Network (CNN). More recent studies use object detection to detect the {\em semantic parts} of the object and extract visual features at the part-level \cite{elhoseiny2017link,zhu2018generative,zhang2016spda}. This
approach makes the image more compatible with the text, as it enables text-terms such as \enquote{crest} to be linked to the visual representation of parts like \enquote{head}. 
\par

The auxiliary information provided to ZSL tasks may be of various kinds, ranging from pre-defined semantic attributes \cite{lampert2009learning,changpinyo2020classifier,atzmon2018probabilistic}, to captions \cite{Xian_2018_CVPR,Sariyildiz_2019_CVPR} to Wikipedia article describing the species \cite{elhoseiny2017link}.
Here we assume the latter scenario. ZSL studies that rely on Wikipedia articles as auxiliary information improve the visual representation and the compatibility function, and use text representations such as Bag-of-Words and TF-IDF, without further text processing. \cite{lei2015predicting,elhoseiny2013write,elhoseiny2016write,elhoseiny2017link,zhu2018generative}. \citet{qiao2016less} used a simple BOW and a L$_{1,2}$-norm objective to suppress the noisy signal in the text. However, this basic treatment of the text is problematic, as it misses crucial information for detecting the correct class.

Recent studies \citep{lu2019vilbert,tan2019lxmert} have shown improved performance on multiple vision-and-language tasks using pre-trained BERT-based models that jointly learn a representation for vision and language. However, they are tuned on relatively short texts and are not optimal for classifying long textual descriptions. \par
In this work, we proceed in a different, yet complementary, direction to previous work, aiming to purposefully model the contribution of the textual modality to ZSL.
We aim to establish the importance of adequately processing the text into a sound representation of visually salient features, in order to increase the vision-and-language compatibility, which can then be effectively learned in an end-to-end manner.

\section{Strong Baseline Model}
\label{task}

\begin{figure*}[th]
\centering
\scalebox{0.9}{
\includegraphics[width=\textwidth]{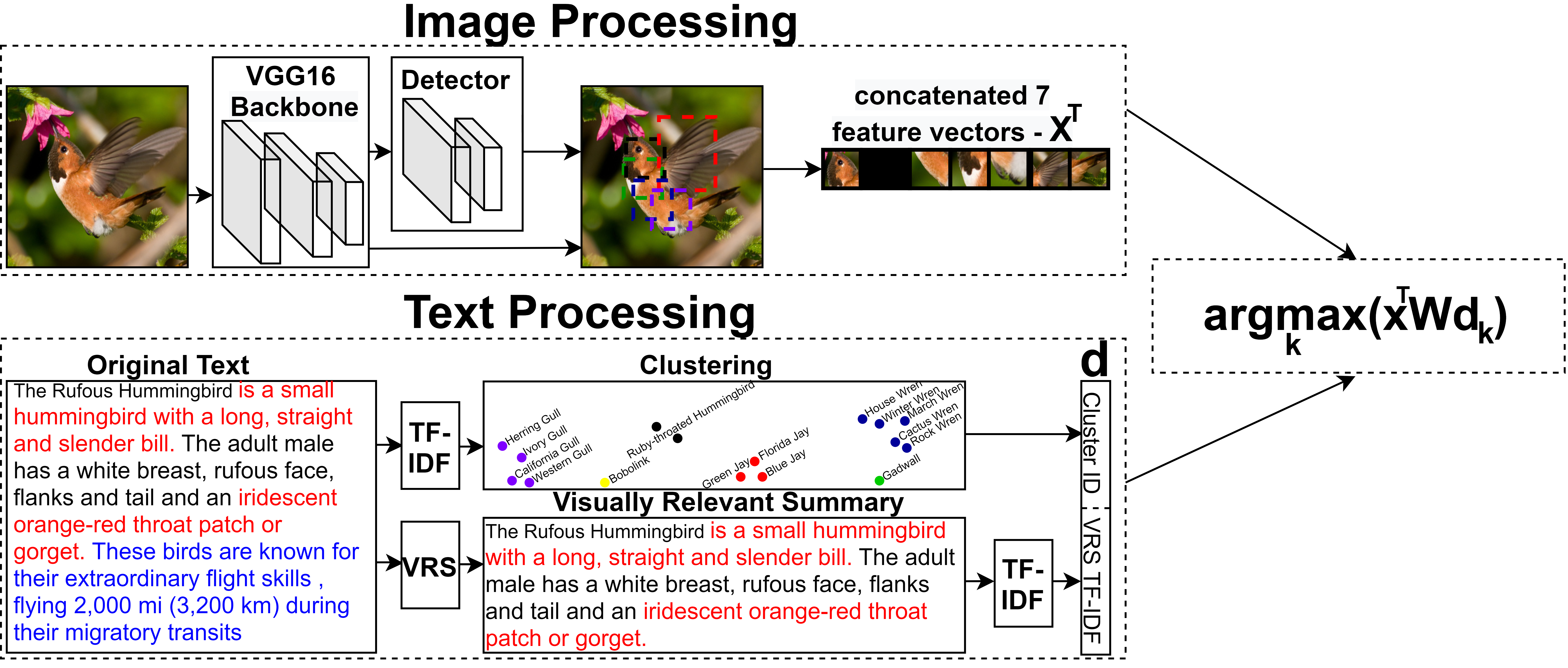}}
 \caption{Our ZEST$_{similarity}$+VRS model with the similarity component and Visually Relevant Summaries (VRS). }
\label{fig:model}
\end{figure*}

The basic architecture, which term ZEST$_{vanilla}$, is a simple multiplicative attention mechanism \cite{luong2015effective} inspired by \citet{romera2015embarrassingly}. We model the problem using an attention-based model, where the image is queried against a set of candidate documents.\par

Formally, let $x^{\mathcal{S}}_1,\ldots,x^{\mathcal{S}}_{M}$ be image feature vectors from a training-set, where $x^{\mathcal{S}}_i\in \mathbb{R}^{m}$. The set of $M$ training images corresponds to a set of $L$ seen classes. Each class has a \textit{single} \enquote{class description} which is a document written by experts in free language (e.g. Wikipedia). We denote $d^{\mathcal{S}}_1,\ldots,d^{\mathcal{S}}_{L}$ as a set of $L$ document feature vectors, 
where $d^{\mathcal{S}}_i\in \mathbb{R}^{\hat{m}}$.
Likewise, let $x^{\mathcal{U}}_1,\ldots,x^{\mathcal{U}}_{N}$ be the image feature vectors from a test set, where $x^{\mathcal{U}}_i\in \mathbb{R}^{m}$. The set of test images corresponds to a set of $K$ unseen classes. Likewise, each class has a \textit{single} \enquote{class description}. We denote $d^{\mathcal{U}}_1,\ldots,d^{\mathcal{U}}_{K}$ as a sets of document feature vectors, 
where $x^{\mathcal{U}}_i\in \mathbb{R}^{\hat{m}}$.
Finally, \(W\in \mathbb{R}^{m\times\hat{m}}\) is our learned matrix.
At inference, the label assignment of
an image $x_i^{\mathcal{U}}$ is defined as:
\begin{equation}
  \label{equation:attention}
  {\hat{y}}=\argmax_{k} \left( x_i^{\mathcal{U}} \right) ^{T}Wd^{\mathcal{U}}_{k}, k \in \{1 \dots K\}
\end{equation}
\par

For an image representation $x_i^{\mathcal{S}}$ and a text representation $d^{\mathcal{S}}_j$, an indicator function $I(x_i^{\mathcal{S}},d^{\mathcal{S}}_j)$ outputs 1 if image $x_i^{\mathcal{S}}$ corresponds to the class described by $d^{\mathcal{S}}_j$ and 0 otherwise. 
The matrix $W$ is then learned by minimizing the categorical cross-entropy loss:

\begin{equation}
\begin{split}
  \sum_{j=1}^{L}I(x_i^{\mathcal{S}},d^{\mathcal{S}}_j) 
  &\times\log(\textit{softmax}({x_i^{\mathcal{S}}}^TWd^{\mathcal{S}}_j))
\end{split}
\end{equation}

\paragraph{Image Encoding}
\label{section:Image_Encoder}
The image encoder's goal is to transform the image into a vector representation of the most salient visual features for the classification. 
We adopt the image encoder for text-based ZSL of \citet{zhang2016spda, zhu2018generative,elhoseiny2017link}. It is based on a Fast R-CNN with \citep{girshick2015fast} a VGG16 backbone for object detection to detect seven semantic parts in the CUB dataset:
\enquote{head},\enquote{back},\enquote{belly},\enquote{breast},\enquote{leg},\enquote{wing},\enquote{tail}. 
Each visual part's encoded features are then concatenated into a feature vector that functions as the image representation for the text-based ZSL.

\paragraph{Text Processing}
\label{section:Text_Encoder}
Our basic encoder processes the text into a feature vector. Similar to previous studies, we employ a TF-IDF representation \citep{salton1988term}. 
We preprocess the text to tokenize words, remove stop words, and stem the remaining words. Then, we extract a feature vector using TF-IDF. This processing procedure is similar to the text processing presented by \citet{zhu2018generative}.
The dimensionalities of TF-IDF features for CUB and NAB are 7,551 and 13,217, respectively. 

\section{The Proposed Approach}
\label{sections:ZEST}
Our solution's key idea is to replace the general class's text representation with a text representation focusing on the most salient features for the visual recognition task.
To do so, we employ two different (complementary) methods: (i) induce a similarity measure used for clustering; and (ii) extract visually relevant text descriptions. Both methods are incorporated in our proposed end-to-end vision-and-language classification architecture, presented in Figure \ref{fig:model}. In what follows we describe the similarity component \ref{section:similarity_component}, and the extraction of visually relevant summaries \ref{section:vrs_component}.

\subsection{The Importance of being Similar}
\label{section:similarity_component}

Our proposed method leverages the similarities between images and texts. That is, when the images look similar, the texts describing their classes are also similar, and vice versa. Here, we propose
to reconstruct this similarity link. 
\par

 To this end, we propose two models: (1) a strong baseline based on two nearest neighbors, which create a link between images and texts; (2) adding a similarity component to our model ZEST$_{vanilla}$.
For both models we use the Image Encoder (section \ref{section:Image_Encoder}) to process the images $x$, and the Text Processing (section \ref{section:Text_Encoder}) to process documents D.

\begin{figure}[t]
\centering
\scalebox{0.49}{
\includegraphics[width=\textwidth]{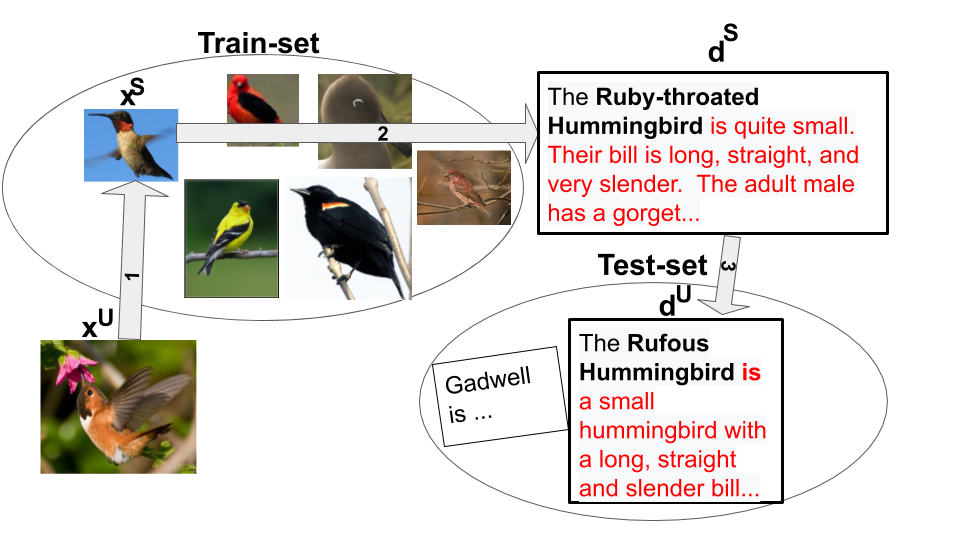}}
 \caption{The Nearest Neighbor Similarity (NNS) model links images and texts through in-modality similarities}
\label{fig:DS}
\end{figure}

\subsubsection{Nearest Neighbor Similarity (NNS)}

Figure \ref{fig:DS} presents our Nearest Neighbor Similarity (NNS) method, which aims to reconstruct the parallel similarity links between the vision and text latent spaces. 
\par

The algorithm is as follows. Given an image $x^{\mathcal{U}}$ from an unseen class in the zero-shot phase,
we first look for the nearest neighbor image in the set of training images, using cosine similarity. The closest image from the training set $x^{\mathcal{S}}_k$ corresponds to a document from training $d^{\mathcal{S}}_{\hat{k}}$. We then look for the nearest neighbor text in from test set $d^{\mathcal{U}}_y$ and predict the corresponding class label y.

\subsubsection{ZEST$_{similarity}$}

A different way to incorporate textual similarity into the classification is to embed it into our model ZEST$_{vanilla}$, to benefit from it in the learning procedure. To this aim, we want to add on top of our text feature vector a representation of the text's similarity to its neighbors. \par

The Basic Encoder captures similarities and differences at the word-level. However, to find similarities at the document level we add to this vector our $similarity$ component, which applies unsupervised clustering to all class descriptions in the training and test texts. We use two different clustering methods that capture different aspects of text similarities. The cluster indexes are then embedded as a BOW (hence cluster embedding). \par

We hypothesize that the similarity component will work well on the \enquote{easy} scenario - where closely related birds are seen during training, and their text can cluster together to indicate these similarities.

\subsection{The Importance of Being Seen}
\label{section:vrs_component}

Here we extract visually relevant features \textit{from the text}, making the texts that enter the classification more compatible with the salient visual information typically reflected in images. 

While the similarity method takes advantage of the similarity between objects seen in training and objects seen at test time, here we want to address the harder scenario, where similar objects are observed together during test time only (e.g. zebras and mules), and they may be very different from those observed during training. 

To differentiate between classes in the test set we need to emphasize the parts that are different, both in the image and the text --- and these are typically their most salient visual features. 

\subsubsection{Visually Relevant Summaries (VRS)}

Our method for enhancing the textual description is based on visually relevant extractive summaries.
Extractive summarization is the task of extracting a small number of sentences that summarize a given document.
In this work, we define {\em visually relevant extractive summarization} (VRS) as the task of extracting only sentences that represent {visually relevant} language. The term {\em visually relevant language} (VRL) was coined by \citet{winn2016detecting} to indicate sentences which are visually descriptive with respect to the object (i.e., bird species).

A {na\"{i}ve} approach for VRS would be to extract sentences with parts that we know are visually salient in our domain (e.g., the 7 parts employed by the vision recognition representation). However, this na\"{i}ve approach has several drawbacks. First, bird parts can be described using many different terms and paraphrases; additionally, a bird can be described by its property values (e.g., black), without any mention of the attribute (e.g., beak). 
Instead, we propose to use the similarity of sentences in the documents and compare them to naturally occurring sentences (`in the wild') containing VRL. \par

Note that we cannot rely on descriptions of particular species due to the zero-shot setup. We must do with descriptions of objects in the {\em general domain} of objects we are interested in classifying. 

\subsubsection{ZEST$_{similarity}$+VRS}
One way to obtain naturally-occurring descriptions of birds is from captions that describe bird images. Critically, these captions need {\em not} be from our dataset, they can describe any bird image. 

We propose to use a set of L bird captions to create an unsupervised classifier. The classifier will receive a set of sentences (assembled as a document), and for each sentence, the classifier will predict whether the sentence is relevant, that is, whether it contains descriptions that can be {\em seen} in a bird image.
\par

For each document, we propose to calculate the pairwise similarity between captions and sentences in the Wikipedia description, and based on this similarity, assign a VRS-score to each sentence.
\par

We calculate the VRS-score of a sentence $s_j$ to a caption by computing the cosine similarity of the embeddings of both the captions ($c_{0:L}$) and sentences ($s_{0:M}$) in the document. For a fixed-size sentence embeddings, we use a pre-trained siamese-and-triplet network \citep{reimers2019sentence, Schroff_2015_CVPR} on top of a pre-trained BERT network \cite{devlin2019bert}.
\par

The VRS-score of sentence $s_j$ with respect to all available captions  $c_{1:L}$ is thus defined to be:
\begin{equation}
\begin{aligned}
\begin{split}
\label{eq:score}
&score(s_j)= \frac{1}{L}
&\sum_{i=1}^{L} \frac{c_i\cdot s_j}{\| s_j\|\| c_i\| }
\end{split}
\end{aligned}
\end{equation}

We then take the highest \textit{k} scoring sentences from \(s_{0:K}\) to be the visually relevant extractive summary of the document. We can then concatenate the similarity embedding to the VRS summary of the text, and perform the multiplicative attention on this revised encoding of the documents and the same image encoding as before.\par

A bird's eye overview of our overall architecture is presented in Figure~\ref{fig:model}.
The text that enters the $similarity$ (clustering) component is the original Wikipedia document, {\em not} the document's VRS summary. Documents contain many non-visual descriptions that are unobserved in the images. However, these non-visual descriptions might still be essential to capture the similarity between documents. For example, similar-looking birds are likely to be in the same habitat. Thus, the VRL sentence extraction and the similarity enhancement operate in parallel on the original document. 
\section{Experiments}

\subsection{Experiment setting}

\paragraph{Datasets:} We evaluate our method \footnote{Our code can be found at \href{https://github.com/tzuf/ZEST}{https://github.com/tzuf/ZEST}.} on the Caltech UCSD Birds-2011 dataset (CUB) \citet{wah2011caltech} and the North America’s birds dataset (NAB) \cite{van2015building}, using class descriptions obtained from Wikipedia and the AllaboutBirds website \footnote{\url{https://dl.allaboutbirds.org}}, collected by \citet{elhoseiny2017link}. 
Both are fine-grained datasets of birds but from different species. The CUB dataset contains 11,788 images of 200 bird species, and the NAB is a larger dataset of birds with 48,562 images of 404 classes\footnote{\citet{elhoseiny2017link} merged the original 1,011 classes according to the subtle division of classes.}.
The texts of both CUB and NAB are long, containing non-visual information. CUB has an average of 869 tokens and 42 sentences in class documents. NAB has an average of 1277 tokens and 58 sentences in class documents. 

\paragraph{Two split Settings} We use the two splits presented by \citet{elhoseiny2017link}: (1) Super Category-Shared (SCS), also referred to as the \enquote*{easy} split; and (2) Super-Category-Exclusive (SCE), also referred to as the \enquote*{hard} split. In the SCS, for each class in the test set, at least one class in the training set belongs to the same category (categories are organized taxonomically). For example, in Figure \ref{fig:bird_example}, the Rufous Hummingbird and the Ruby-throated Hummingbird are both from the Hummingbird category. In the SCE, all classes in a category are in the same set. Namely, if a class is in the test set, then other classes from the same category are also in the test set, and will never be seen during training. Intuitively, classes from the same category have high similarity in both images and texts, so while in SCS similar images have been seen during training, in the SCE a class from an entirely new category is seen for the first time. 

\paragraph{Training Details:}
The parameters of our model include cluster parameters. We use two clustering methods: (1) Density-based spatial clustering of applications with noise (DBSCAN) \citep{ester1996density}; (2) Hierarchical DBSCAN \citep{mcinnes2017hdbscan}. The DBSCAN algorithm takes two parameters: (1) \enquote{minimal cluster} - the number of samples in a neighborhood for a point to be considered as a core point; (2) \enquote{max distance} the maximum distance between two samples for one to be considered as in the neighborhood of the other. 
The \enquote{minimal cluster} is chosen to be two as two birds are the minimal similarity we want (similar to the NNS model).
The \enquote{max distance} parameter we optimize on validation sets (10\% of data) according to the two splits. In addition, the similarity model includes a threshold for performing the similarity component, also optimized over the validation set. 
The VRS algorithm includes a sentence score threshold for the number of sentences to be extracted. This threshold was chosen on the validation set.

The weights   $W$ were initialized with normalized initialization \citep{glorot2010understanding}. The cross-entropy loss function was optimized with Adam optimizer \citep{Adam}.

\paragraph{Human Summarization:} 
To evaluate our proposed VRS extraction method, we designed an oracle experiment using ground-truth visually relevant summarization. To this end, two independent human experts manually annotated the CUB dataset by reading each sentence in the document and marking the sentence as yes\textbackslash no VRL. We set guidelines to resolve disagreements (e.g. hatchlings descriptions were marked as not VRL). On average, only 11.9\% of the sentences were found to include VRL.

\paragraph{Image Captions:}
To create visual summaries we use image captions of birds from the CUB train set, provided by \citet{reed2016learning}. Each image in the CUB dataset has been annotated with ten fine-grained captions. These captions describe only the birds' visual appearance while avoiding mentioning the names of the bird species. E.g., \enquote{This bird has a long beak, a creamy breast, and body, with brown wings}. In this work, we use the first five captions of each image.

To showcase this approach's generality, we use these captions in both in-domain (CUB) and out-of-domain (NAB) scenarios. In all cases, we avoid using captions of unseen (test) bird classes. In NAB, we effectively use captions from CUB to extract VRS for entirely-different species presented in NAB. Note that {\em only} models that include the VRS component (+VRS) employ these image captions. We report the accuracy achieved per the number of captions used in the VRS, to indicate the number of captions that are realistically needed.

\paragraph{Baselines:} 
Our approach is compared asainst ten leading algorithms (see Table \ref{tab:results}): MCZSL \citep{akata2016multi}, WAC-Linear \citep{elhoseiny2013write}, Wac-Kernel  \citep{elhoseiny2016write}, ESZSL   \citep{romera2015embarrassingly}, SJE  \citep{akata2015evaluation}, Sync$_{fast}$ \citep{changpinyo2016synthesized}, Sync$_{OVO}$ \citep{changpinyo2016synthesized},
ZSLNS \citep{qiao2016less}, and GAZSL \citep{zhu2018generative}.

\paragraph{Generalized Zero-Shot Learning:} 
The conventional zero-shot learning task considers only unseen classes during the zero-shot phase. However, in a realistic scenario, seen objects might also appear \cite{chao2016empirical}. In Generalized Zero-Shot Learning (GZSL), test data might also come from seen classes, and the labeling space is the union of both types of seen and unseen classes. GZSL is thus considered a more challenging problem setting
than ZSL due to the model's bias towards the seen classes.
We follow the metric present by \citet{chao2016empirical} to evaluate our models on the GZSL task. We evaluate the accuracy of a Seen-Unseen accuracy Curve (SUC) and use Area Under SUC to measure the general capability of ZSL methods.

 \begin{table}[t]
 \centering
 \scalebox{0.6}{

\begin{tabular}{l|ll|ll} 
\Xhline{6\arrayrulewidth}
\multirow{2}{*}{\textbf{ methods} }  & \multicolumn{2}{c|}{\textbf{CUB } }   & \multicolumn{2}{c}{\textbf{NAB} } \\ 
\cline{2-5}
  & \multicolumn{1}{c}{\textbf{SCS} } & \multicolumn{1}{c|}{\textbf{SCE} } & \multicolumn{1}{c}{\textbf{SCS} } & \multicolumn{1}{c}{\textbf{SCE} } \\ 
\hline
MCZSL \citet{akata2016multi}   & 34.7   & -  & -  & -  \\
WAC-Linear \citet{elhoseiny2013write}  & 27.0   & 5.0 & -  & -  \\
WAC-Kernel \citet{elhoseiny2016write} & 33.5   & 7.7 & 11.4   & 6.0 \\
ESZSL \citet{romera2015embarrassingly} & 28.5   & 7.4 & 24.3   & 6.3 \\
SJE \citet{akata2015evaluation} & 29.9   & -  & -  & -  \\
ZSLNS \citet{qiao2016less} & 29.1   & 7.3 & 24.5   & 6.8 \\
SynC$_{fast}$ \citet{changpinyo2016synthesized}   & 28.0   & 8.6 & 18.4   & 3.8 \\
SynC$_{OVO}$ \citet{changpinyo2016synthesized}   & 12.5   & 5.9 & -  & -  \\
ZSLPP \citet{elhoseiny2017link} & 37.2   & 9.7 & 30.3   & 8.1 \\
GAZSL \citet{zhu2018generative} & 43.7   & 10.3 & 35.6   & 8.6 \\ 
\hline
Nearest Neighbor Similarity (NNS) & 40.402  & 5.551   & 37.002  & 5.517   \\
ZEST$_{vanilla}$   & 39.16  & 11.77   & 27.61  & 10.18   \\
\hline
\textbf{ZEST$_{similarity}$} & \textbf{47.48} & \textbf{11.77}  & \textbf{38.2} & \textbf{10.18} \\
\hline
\textbf{ZEST$_{similarity}$+VRS}  & \textbf{48.57} & \textbf{15.26} & \textbf{38.51 } & \textbf{10.23 } \\ 
\Xhline{6\arrayrulewidth}
   
\end{tabular}
}
\caption{Top-1 accuracy (\%) on CUB and NAB datasets with two split settings. We report the mean over three random initializations. The standard-deviation for  ZEST$_{similarity}$ for the CUB is 0.337 and 0.368; for NAB 0.625 and 0.174 (for the SCS and SCE splits accordingly). }
\label{tab:results}
\end{table}

\begin{table}[t]
\centering
 \scalebox{0.61}{
\begin{tabular}{l|cc|cc} 
\Xhline{6\arrayrulewidth}
\multirow{2}{*}{\textbf{ methods} } & \multicolumn{2}{c|}{\textbf{CUB } } & \multicolumn{2}{c}{\textbf{NAB} }   \\ 
\cline{2-5}
   & \textbf{SCS}   & \textbf{SCE}   & \textbf{SCS}  & \textbf{SCE}  \\ 
\hline
GAZSL & \begin{tabular}[c]{@{}c@{}}43.74\\\end{tabular} & \begin{tabular}[c]{@{}c@{}}10.3\\\end{tabular} & 35.6   & 8.6 \\
GAZSL+parts summarization & \begin{tabular}[c]{@{}c@{}}19.54\\\end{tabular} & \begin{tabular}[c]{@{}c@{}}9.557\\\end{tabular} & 23.32
   & 7.2
 \\
GAZSL+parts summarization+${similarity}$ & \begin{tabular}[c]{@{}c@{}}38.25\\\end{tabular} & \begin{tabular}[c]{@{}c@{}}9.557\\\end{tabular} & 33.05
   & 7.2
 \\
GAZSL+our VRS  & 43.72 & \textbf{16.44}  & 37.28
  & 9.237   \\
GAZSL+HUMAN & 35.98 & 21.81   & - & -  \\
GAZSL+HUMAN+${similarity}$   & 47.32 & 21.81   & - & -  \\ 
\hline
   ZEST$_{vanilla}$   & 39.16  & 11.77   & 27.61  & 10.18   \\

 ZEST$_{vanilla}$+our VRS   & {42.58}  & {15.26}  & {32.24}   & {10.23} \\
ZEST$_{similarity}$ & 47.48 & 11.77   & 38.2
  & 10.18   \\
ZEST$_{similarity}$+parts summarization & 42.27
 & 10.93
   & 37.02

  & 8.055
   \\
ZEST$_{similarity}$+our VRS   & \textbf{48.57}  & {15.26}  & \textbf{38.51}   & \textbf{10.23} \\

ZEST$_{similarity}$+HUMAN & 48.99 & 17.2 & - & -  \\ 
\Xhline{6\arrayrulewidth}
\end{tabular}
}
  \caption{Visually Relevant Summarization (VRS) with GAZSL, ZEST$_{vanilla}$, and ZEST$_{similarity}$. }%
  \label{table:summarization}%
\end{table}

\begin{table}[t]
\centering
 \scalebox{0.87}{
\begin{tabular}{l|c|c} 
\Xhline{6\arrayrulewidth}
  \multirow{2}{*}{ \textbf{methods }} & \textbf{CUB } & \textbf{NAB }  \\ 
\cline{2-3}
   & \textbf{SCS } & \textbf{SCS}  \\ 
\hline
ZEST$_{vanilla}$  & 39.16  & 27.61  \\
ZEST$_{vanilla}$+bird category & 43.71  & 36.73   \\
Zest$_{similarity}$ only 1 cluster & 46.55  & 35.94   \\
Zest$_{similarity}$ full (2 cluster) & \textbf{47.48} & \textbf{38.2}  \\
\Xhline{6\arrayrulewidth}
\end{tabular}
}
  \caption{Zest model with different similarity methods}%
  \label{table:similarity}%
\end{table}

\begin{table}[t]
\centering
 \scalebox{0.76}{
\begin{tabular}{l|cc|cc} 
\Xhline{6\arrayrulewidth}
   \multirow{2}{*}{\textbf{ methods} } & \multicolumn{2}{c|}{\textbf{CUB } } & \multicolumn{2}{c}{\textbf{NAB} }   \\ 
\cline{2-5}
   & \textbf{SCS}   & \textbf{SCE}   & \textbf{SCS}  & \textbf{SCE}  \\ 
\hline
ESZSL & 0.185   & 0.045 & 0.092   & 0.029 \\
ZSLNS & 0.147   & 0.044 & 0.093   & 0.023 \\
WAC$_{kernal}$  & 0.225   & 0.054 & 0.007   & 0.023 \\
WAC$_{linear}$  & 0.239   & 0.049 & 0.235   & -   \\
SynC$_{fast}$  & 0.131   & 0.040 & 0.027   & 0.008 \\
SynC$_{OvO}$   & 0.017   & 0.010 & 0.001   & -   \\
ZSLPP & 0.304   & 0.061 & 0.126   & 0.035 \\
GAZSL & 0.354   & 0.087 & 0.204   & 0.058 \\ 
\hline
ZEST$_{similarity}$ & \textbf{0.443 } & \textbf{0.1 }   & \textbf{0.267 } & \textbf{0.067 }  \\
\hline
ZEST$_{similarity}$+VRS   & \textbf{0.437}   & \textbf{0.147} & \textbf{0.26}   & \textbf{0.084} \\
ZEST$_{similarity}$+HUMAN & 0.445   & 0.163 & -  & -   \\
\Xhline{6\arrayrulewidth}
\end{tabular}
}
  \caption{Generalized Zero-Shot Learning: AUC of Seen-Unseen Curve.}%
  \label{table:GZSL}%
\end{table}

\subsection{Results}
Table \ref{tab:results} presents the top-1 accuracy for each of the models. 
The table is divided into four sections, which are (from top to bottom): 
(1) previous work; (2) our baselines ;(3) our models with previous setup (for comparison to previous work); (4) Our model with additional data - captions. 

\paragraph{NNS Model:} 
According to Table \ref{tab:results}, the NNS model achieves competitive results on the SCS - 40.402\% and 37.002\% on CUB and NAB correspondingly. The high scores on the SCS, where similar birds have been seen during training, is expected --- as this method relies on similarities within texts and images. 
In contrast, the NNS suffers from low accuracy on the SCE, where different categories of birds have been seen during training. As the NNS model relies on text and image similarities, it is intuitively appealing that low accuracy on the SCE stems from the fact that birds from different categories are less likely to look alike. 

\paragraph{ZEST$_{vanilla}$}
In contrast to the very sophisticated approaches of \citet{zhu2018generative}, the vanilla cross-entropy based approach outperforms all previous methods on the SCE-split on both CUB (+14.27\% ratio of improvement) and NAB (+18.37\% ratio of improvement). As the SCE-split is a more challenging split, this sheds light on the strength as well as limitations of this simple framework. 

\paragraph{ZEST$_{similarity}$}

We then combined strengths of ZEST$_{vanilla}$ and NNS models over the two different scenarios: \enquote{hard} and \enquote{easy}, respectively. \par

\begin{table*}
\centering
 \scalebox{0.85}{
\begin{tabular}{ll||l||ll} 
\Xhline{6\arrayrulewidth}

&Sentence                                                                                                                                                                                                                                & HUMAN & VRS Model  \\ 
\Xhline{6\arrayrulewidth}
1 & \begin{tabular}[c]{@{}l@{}}After nesting, north american birds move in flocks further north along the coasts, \\returning to warmer waters for winter.\end{tabular}                                                                      & \multicolumn{1}{c||}{\xmark}     & \multicolumn{1}{c}{\xmark}         \\ 
\hline
2 & \begin{tabular}[c]{@{}l@{}}Red foxes and coyotes readily predate colonies that they can access, the later \\being the only known species to hunt adult pelicans (which are too large for \\most bird predators to subdue).\end{tabular} & \multicolumn{1}{c||}{\xmark}    & \multicolumn{1}{c}{\xmark}         \\ 
\hline
3 & \begin{tabular}[c]{@{}l@{}}when foraging, they dive bill-first like a kingfisher often submerging completely \\below the surface momentarily as they snap up prey.\end{tabular}                                                         & \multicolumn{1}{c||}{\xmark}    & \multicolumn{1}{c}{\xmark}         \\ 
\hline
4 & It is one of only three pelican species found in the western hemisphere.                                                                                                     & \multicolumn{1}{c||}{\xmark}    & \multicolumn{1}{c}{\cmark}         \\ 
\hline
5 & Due to their small size, they are vulnerable to insect-eating birds and animals.                                                                                                                                                        & \multicolumn{1}{c||}{\xmark}    & \multicolumn{1}{c}{\cmark}         \\ 
\hline
6 & Hummingbirds show a slight preference for red, tubular flowers as a nectar source.                                                                                                                                                      & \multicolumn{1}{c||}{\xmark}    & \multicolumn{1}{c}{\cmark}         \\
\hline

7 & The head is white but often gets a yellowish wash in adult birds.                                                                                                                                                      & \multicolumn{1}{c||}{\cmark}     & \multicolumn{1}{c}{\cmark}         \\
\Xhline{6\arrayrulewidth}
\end{tabular}
}
\caption{Qualitative analysis, showing seven sentences from three randomly selected summaries. The table shows HUMAN and VRS model markings of the sentence as yes( \cmark)\textbackslash no(\xmark) VRL. 
}
\label{tab:qualitative}
\end{table*}

The ZEST$_{similarity}$ model adds the cluster index embedding to the TF-IDF representation, only if a significant percentage of the documents from the test-set are clustered with documents from the train set. The threshold picked over the validation set is a 15\%. 
Thus, in the case of the SCE-split, no or few similarities are found, and the ZEST$_{similarity}$ preforms at the same level as the ZEST$_{vanilla}$ model. The threshold parameter was optimized on the validation set. \par

The two clustering algorithms we applied find real similarities, achieving high accuracy when tested on predicting the correct label according to the ground-truth taxonomical category. The HDBSCAN, and DBSCAN achieved 88\% and 84.5\% accuracy on the CUB, and 93.07\% and 95.05\% on the NAB, accordingly.

Interestingly though, different clustering find different sources of similarities, that are essentially {\em additive}. In Table \ref{table:similarity} we can see a comparison between different similarity enhancing methods. 
The ZEST$_{vanilla}$+bird category method is a BOW of the bird category added to the original text embedding and then passed as before to a ZEST$_{vanilla}$ model. 
The use of two clusters that capture different similarities performs better than embedding the bird category in the text representation, by a ratio improvement of up to 8.63\%. This suggests that our $ZEST_{similarity}$ method captures similarities that are beyond the bird category. 
\par

Finally, in Table \ref{table:GZSL} we present the results of ZEST$_{similarity}$ in the GZSL setup. On both datasets and splits, the ZEST$_{similarity}$ achieves state-of-the-art results with up to 30.88\% ratio improvement.

\paragraph{ ZEST$_{vanilla}$+VRS and ZEST$_{similarity}$+VRS }
use the captions from training images in the CUB in order to generate visually relevant extractive summaries of the original Wikipedia documents.

We test the summarized representation on the ZEST$_{vanilla}$ model, the ZEST$_{similarity}$, and the GAZSL \citep{zhu2018generative} model. In Table \ref{table:summarization} we show the experimental results. We compare the models before and after the use of the Visually Relevant Extractive Summarization component. We see an improvement in accuracy in both models on both datasets and on both splits. \par

In contrast to the ZEST$_{similarity}$, the GAZSL does not have a component that embeds similarities. The VRS reduces similarity by removing non-VRL that might be similar between documents. 
The HUMAN summary is an especially lean summary with only 11.9\% sentences extracted. Thus, the similarity between texts of similar objects diminishes. The GAZSL+HUMAN in the SCS-split performs poorly due to the diminished similarity. In contrast, The GAZSL+HUMAN+our VRS adds the similarity that was lost and the performance improves.
\par

To assess the quality of the VRS summarization performance, we treat HUMAN summarization as the ground truth. The VRS method succeeds in removing 49.4\% of the sentences in the CUB dataset with 96.23\% recall and 22.59\% precision. For comparison, removing 49.4\% of the sentences randomly produces a recall of 50.6\% and a precision of 11.9\%. \par

Table \ref{tab:qualitative} shows a qualitative analysis of our VRS results.  In sentences 1-3, the VRS model correctly marked the sentences as non-VRL: sentence 1 is a typical case of non-visually-relevant language describing birds migration; in sentence 2 the VRS model correctly marks the sentence as non-VRL despite the mention of color (red) --- since the color does {\em not} refer to the object to be classified (the bird); in sentence 3 the VRS model correctly marks the sentence as non-VRL despite the mention of a body part (bill) --- since that description it is not visually relevant in that particular context. Sentences 4-6 show examples of false-positive predictions of the VRS model. E.g., in sentence 5 the VRS model incorrectly predicted VRL, which we attribute to the mention:  \enquote{their small size}. In sentence 6 the VRS model incorrectly marks the sentence as VRL, a mistake we attribute to the mention of the flower's \enquote{red} color. \par

We then compare both ZEST$_{similarity}$ and the GAZSL to the use of HUMAN summarization in the CUB dataset and see additional improvement in both models on the two splits. The gap between the performance on the VRS and the Human summarization indicates that improvement in the summarization of documents will improve the models' performance, and is, therefore, a promising path for text-based zero-shot learning research.

\begin{figure}[t]
\centering
\scalebox{0.47}{
\includegraphics[width=\textwidth]{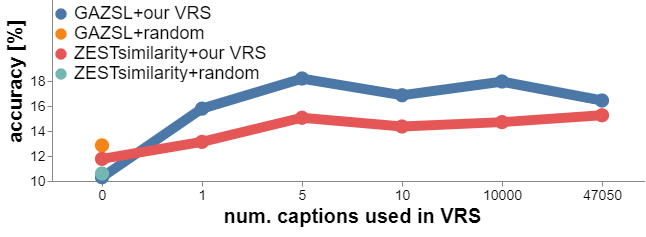}}
 \caption{Accuracy per number of captions used to \textit{focus}  summarization, measured on the hard SCE split of CUB. Showing that as little as 5 captions in total are sufficient to focus the summarization process.
 }
\label{fig:captions}
\end{figure}

Finally, we experiment to assess the number of captions that are realistically needed for the VRS method. The results, presented in Table \ref{fig:captions}, show that only a few ($\sim5$) 
 sentences (captions) from arbitrary birds are needed to achieve the maximum accuracy with this method.
Testing the VRS with five arbitrary captions from CUB dataset on the NAB dataset with SCS-split, we achieved a 39.28\% accuracy. \par

For comparison, \citet{reed2016learning} showed that their model needed at least 512 captions per class to achieve the maximum accuracy - i.e., had it used all the captions available.

\par

\section{Conclusion}
\label{conclusion}

This work aims to establish a better way to represent the language modality in {\em text-based ZSL for image classification}. Our approach only relies on semantic information about visual features, and not on the visual features themselves.
Specifically, our two orthogonal text-processing methods, employing {\em textual similarity} and {\em visually-relevant summaries}, lead to significant improvements across models, splits, and datasets, and illustrate that adequate text-processing is essential in text-based ZSL tasks.  We   conjecture that text-processing methods will be essential in a range of vision and language-based tasks, and hope this work will assist future research in better representing the language modality in various multi-modal tasks.


\section{Acknowledgments}
The research of the first and last author is funded by the European Research Council (ERC grant \#677352) and the Israel Science Foundation (ISF grant \#1739/26), and the research of the third author is also funded by the Israel Science Foundation (ISF grant \#737/2018), for which we are grateful.

\bibliography{emnlp2020}
\bibliographystyle{acl_natbib}
\end{document}